\documentclass[screen]{acmart}
\usepackage{balance}

\newcommand{\tool}{EarlyCommit}

\begin{document}

\title{EarlyCommit: Calibrated Word-by-Word Commitment for Robots Acting Before You Finish Speaking}


\author{Muhammad Haris Khan}
\affiliation{%
  \institution{University of Copenhagen}
  \city{Copenhagen}
  \country{Denmark}}
\email{dcg967@ku.dk}

\renewcommand{\shortauthors}{Khan et al.}

\begin{abstract}
    Human listeners can often identify the intended object several words before an instruction ends. Robots, however, are typically designed to wait until the speaker has finished before deciding how to act. 
    We present \tool, an open-source tool that lets a robot commit to the intended object while the instruction is being spoken. 
    After every recognised word a fine-tuned text encoder scores each object in the scene. The robot acts when the calibrated probability of the best object reaches a threshold $\tau$ set by the user, so $1-\tau$ is the accepted risk of choosing the wrong object. 
    Otherwise it waits or asks. 
    The tool includes a streaming speech front end, a scene and instruction generator with exact word-by-word ground truth, and scripts to train, calibrate and evaluate. On instructions containing words never seen in training, \tool{} commits correctly 1.6 to 2 times as often as keyword matching, is right at least 92\% of the time when it commits at $\tau=0.9$, and needs 31--35\,ms per word on consumer GPUs.
\end{abstract}



\begin{teaserfigure}
     \centering
      \includegraphics[width=0.93\textwidth]{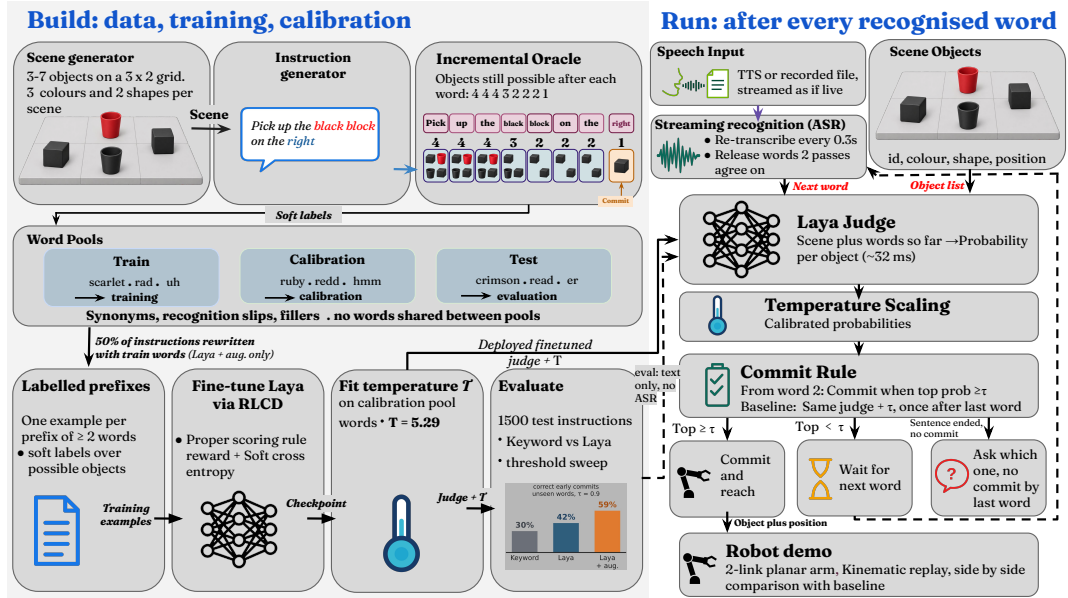}
      \caption{\tool{} architecture. Left: the build pipeline. Generated scenes and instructions are labelled after every word by an incremental oracle. Three word pools with no shared words supply unseen language: the train pool for augmentation, the calibration pool for fitting the temperature and the test pool for evaluation. Right: the loop that runs after every recognised word. The judge scores each scene object, the probabilities are calibrated, and the robot commits, waits for the next word or asks which object is meant. The evaluation feeds generated words to the judge directly, without speech recognition.}
      \Description{Two-panel diagram. The left panel shows the offline pipeline from scene and instruction generation, through an incremental oracle and three disjoint word pools, to fine-tuning, temperature fitting and evaluation. The right panel shows the runtime loop: speech is transcribed word by word, a judge scores each scene object, probabilities are calibrated, and a commit rule chooses between committing, waiting for the next word, or asking. Arrows connect the calibrated judge from the left panel into the runtime loop.}
      \label{fig:arch}
\end{teaserfigure}


\maketitle

\section{Background and Summary}

A person asks a robot: "could you grab the red cup, the one on the left next to the laptop, and bring it over here, please". If only one red cup stands on the table, a human helper knows the answer at the word "cup", the sixth of 21 words, and starts to reach while the remaining 15 words, more than four seconds of speech, are still being said. A typical robot pipeline instead waits for the end of the utterance, transcribes it and then decides. This adds delay to every request, and the delay grows with every word the speaker adds.
Users prefer a robot to reply about one second after their request~\cite{shiwa2008quickly}, and robots that act in anticipation finish shared tasks seconds faster than robots that only react~\cite{huang2016anticipatory}.
Although acting early is useful, it is also risky. 
For example, halfway through a sentence several objects can still fit, and a wrong grasp costs more than a short wait. 
A robot that acts early therefore needs two things. 
It must know after each word which object is meant, and it must know when that belief is reliable enough to move.

Prior work covers each need separately. 
Incremental reference resolution scores scene objects word by word~\cite{schlangen2009incremental,kennington2015simple}, learns when a partial utterance is understood well enough to respond~\cite{devault2009finish}, and lets robots act on partial utterances~\cite{brick2007incremental} or show their uncertainty through their actions~\cite{hough2017grounding}. 
These systems do not tie the decision to act to a calibrated error rate, and their word-level models do not transfer to words outside their training vocabulary. Conformal methods let robots ask for help with statistical guarantees~\cite{ren2023robots}, and risk-controlled rules stop classification early on growing inputs~\cite{ringel2024early}. 
They work on complete instructions or on fixed label sets. 
Concurrent work lets a vision-language-action policy act on partial instructions through a learned readiness threshold~\cite{park2026premover}. Speech research on turn-taking answers a different question~\cite{SKANTZE2021101178}. Its silence-based, inter-pausal-unit and continuous models decide \emph{when} the user's turn ends so that the system can speak or give feedback, and incremental dialogue systems pass partial results between modules as words arrive. \tool{} uses such an incremental front end but asks \emph{what} is meant: which object, and whether it is safe to move for it while the speaker is still talking. It never predicts the end of the turn, and its decision is an action whose error rate the user sets with $\tau$. A recent survey finds very little research on this kind of incremental decision making~\cite{kennington2025prior}.

\tool{} brings these pieces together as one reusable tool. We contribute
(1)~a runtime pipeline that turns streaming speech into a word-by-word decision to commit, wait or ask, controlled by one interpretable threshold $\tau$;
(2)~an evaluation harness that generates scenes and instructions with exact ground truth after every word and tests robustness to unseen language;
(3)~an evaluation showing when the threshold holds and a practical lesson on how to calibrate it.
Code, result files and demo videos are provided as anonymised supplementary material.
\begin{figure*}[t]
  \centering
  \includegraphics[width=0.95\textwidth]{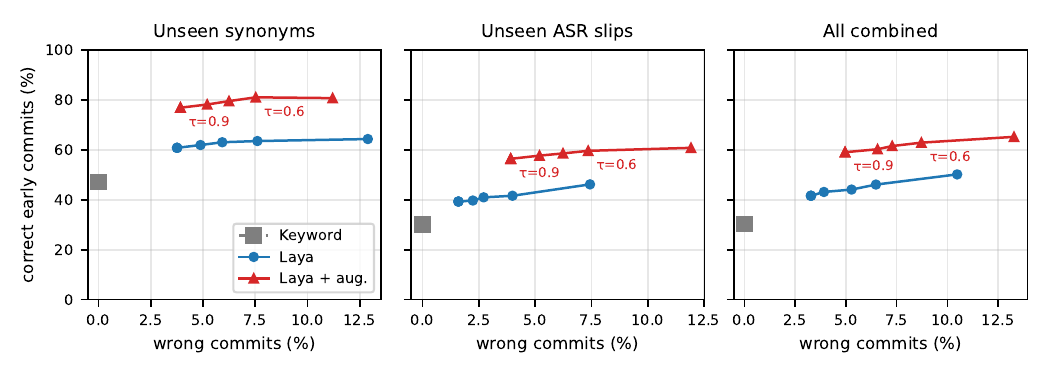}
  \caption{Correct against wrong commits as $\tau$ varies on unseen language. Up and to the left is better. Keyword matching does not depend on $\tau$ and appears as one point. Points for $\tau$ between 0.85 and 0.95 coincide.}
  \Description{Three line plots of correct commits against wrong commits for keyword matching, Laya and Laya with augmentation.}
  \label{fig:tradeoff}
\end{figure*}

\section{Purpose}

\tool{} is a drop-in module for robots that take spoken requests: given the list of objects on the table and the words recognised so far, it returns after every word either an object to grasp, ``wait'', or ``ask''. It also lets researchers train, calibrate and evaluate this decision on their own scenes and vocabularies, with exact word-by-word ground truth and one threshold $\tau$ that sets the accepted risk of a wrong grasp.

\section{Characteristics}

\subsection{Runtime pipeline}
Figure~\ref{fig:arch} (right) shows the modules.

\textbf{Streaming speech recognition.} Whisper~\cite{radford2023whisper} (faster-whisper, \texttt{small.en}) re-transcribes the audio received so far every 0.3\,s. A word is released only when two consecutive transcripts agree on it, following the LocalAgreement policy~\cite{machacek2023turning}. 

\textbf{Judge.} The judge is Laya~\cite{laya2026}, a 421M-parameter text classifier built on the ModernBERT-large encoder~\cite{warner2024modernbert}. It does not generate text. It receives the scene as structured text, for example \texttt{\{id: obj\_2, color: red, shape: cup, position: middle front\}} for each object, and the words heard so far. It also receives one multiple-choice question, ``Which object on the table does the speaker mean, given only the words said so far?'', whose options are the scene objects. Each option is scored at its own marker token and a softmax over the objects gives the probabilities. Options are defined per request, in the spirit of zero-shot label scoring~\cite{yin2019benchmarking}. The judge re-reads the whole prefix after every word, the restart-incremental use of a bidirectional encoder~\cite{madureira2020incremental}.

\textbf{Calibration.} Raw probabilities $p_t$ are rescaled with one temperature $T$~\cite{guo2017calibration}, $\tilde p_t=\mathrm{softmax}(\log p_t/T)$, so that a probability of 0.9 means the object is right about 90\% of the time. $T>1$ softens the distribution and $T<1$ sharpens it.

\textbf{Commit rule.} Let $\tilde p_t$ be the calibrated distribution over objects after word $t$. The robot commits to $\arg\max_k \tilde p_{t,k}$ at the first $t\geq2$ with $\max_k \tilde p_{t,k}\geq\tau$. The first word is skipped because it is always part of the verb. A commit is final. If the instruction ends without a commit the robot asks which object is meant. The full-utterance baseline applies the same judge and $\tau$ once at the end of the instruction.

\textbf{Robot interface and demo.} A commit returns the object identifier and position to the caller. The included demo replays these decisions on a kinematic two-link arm, beside the full-utterance baseline.

\subsection{Evaluation harness}
Figure~\ref{fig:arch} (left) shows how data with exact labels is produced without annotation.

\textbf{Scenes.} Each scene has 3 to 7 objects on a $3\times2$ grid (left, middle, right by front, back). The vocabulary has 6 colours and 5 shapes, but each scene draws only 3 colours and 2 shapes. Objects therefore share attributes and instructions need several words to single one out.

\textbf{Instructions.} A target is chosen and its attributes are selected in the order colour, shape, column, row until no other object fits. With probability 0.2 one attribute is dropped, which can leave the instruction ambiguous. Otherwise, with probability 0.5, a redundant attribute is added. Word order varies, and half of the instructions end with words that carry no information such as ``please'' or ``for me''.

\textbf{Incremental oracle.} After every word the oracle lists the objects still consistent with the attribute words heard so far. This gives a soft label for each prefix (uniform over the consistent objects), the earliest word at which the target is determined and an ambiguity flag. In Figure~\ref{fig:arch} the counts after each word of ``pick up the black block on the right'' are 4, 4, 4, 3, 2, 2, 2, 1.

\textbf{Unseen language.} Instructions can be rewritten with synonyms (red to crimson), speech recognition (ASR) slips into non-attribute words (red to ``read'') and fillers (``er''), keeping the labels exact. Every word list is split into three disjoint pools: one for training, one for calibration and one for testing. A separate stress test uses slips that turn one attribute into another (black to ``back''), which no listener can recover.

\subsection{Training and calibration}
\textbf{Why fine-tune.} Off the shelf Laya picks the right object on only 44\% of validation prefixes whose words already single it out.

\textbf{Training data.} The harness generates 4{,}000 training and 500 validation instructions with different random seeds from the 1{,}500 test instructions. Each prefix of two or more words becomes one example, giving 30{,}972 training examples. An example holds the scene, the prefix, the question and a soft target $y$ that is uniform over the objects the oracle still considers possible. Training on this target teaches the judge to stay uncertain while several objects fit and to become confident exactly when the words decide. The augmented variant rewrites half of the instructions with the training word pool (33{,}007 examples).

\textbf{Objective.} We keep Laya's training recipe, RLCD: reinforcement learning against strictly proper scoring rules~\cite{laya2026}. For each example the logits $z$ over objects are perturbed $G=4$ times with zero-mean Gaussian noise $\epsilon_g$, whose scale $\sigma$ decays from 0.4 to 0.1 over training. Each perturbed distribution $q_g=\mathrm{softmax}(z+\epsilon_g)$ is rewarded with a strictly proper scoring rule $r_g=S(q_g,y)$, a weighted sum of the log and spherical scores. A proper scoring rule is maximised in expectation only by reporting the true probabilities. With advantage $A_g$, the reward minus the group mean divided by the standard deviation, the loss is the policy-gradient term $-\tfrac1G\sum_g A_g\log\mathcal N(z+\epsilon_g;z,\sigma^2 I)$ plus soft cross-entropy $-\sum_k y_k\log\mathrm{softmax}(z)_k$.

\textbf{Optimisation.} We train the whole model, encoder and decision head, for 2 epochs with AdamW (learning rate $2.5\times10^{-5}$ for the encoder and $10^{-4}$ for the head, weight decay 0.01, cosine schedule), batch size 64, gradient clipping at 1 and bfloat16 on one A100 GPU.

\textbf{Temperature fitting.} $T$ minimises the soft-label negative log-likelihood $-\sum_i\sum_k y_{ik}\log\tilde p_{ik}$ over all prefixes of the 500 validation instructions, searched over $\log T\in(-3,3)$. Half of these instructions are rewritten with the calibration pool, whose words appear neither in training nor in testing. This gives $T=5.29$ for both models. $T$ is applied on top of the internal temperature that Laya's training script stores in the checkpoint.

\subsection{Behaviour}
We evaluate on 1{,}500 test instructions per condition (1{,}250 with a unique target, 250 ambiguous). The evaluation feeds the generated words directly to the judge, without speech recognition, so the results isolate the judge and the commit rule. We compare keyword matching, which is the oracle's own parser and is perfect on the templates, with Laya and Laya trained with augmentation, the default judge of \tool. A \emph{correct commit} is a commit to the target on an instruction with a unique target. A \emph{wrong commit} is a commit to any other object. Every commit on an ambiguous instruction is a \emph{false commit}. \emph{Accuracy at commit} is the share of commits that hit the target. \emph{Time saved} counts the words spoken after the commit at 0.3\,s per word. \emph{Calibration error} is the 10-bin expected calibration error of the top probability over all prefixes. Table~\ref{tab:main} reports $\tau=0.9$. 
\begin{table}[t]
  \caption{Correct / wrong commits (\% of instructions with a unique target) at $\tau=0.9$. Unseen conditions use test-pool words only.}
  \label{tab:main}
  \small
  \begin{tabular}{@{}lccc@{}}
    \toprule
    Condition & Keyword & Laya & Laya + aug. \\
    \midrule
    Clean (in-distribution) & 100 / 0 & 100 / 0 & 99.8 / 0 \\
    Unseen disfluencies & 100 / 0 & 99.7 / 0.3 & 99.3 / 0 \\
    Unseen synonyms & 47.1 / 0 & 60.8 / 3.8 & 76.9 / 3.9 \\
    Unseen ASR slips & 30.2 / 0 & 39.3 / 1.6 & 56.4 / 3.9 \\
    All combined & 30.2 / 0 & 41.6 / 3.3 & 59.0 / 5.0 \\
    Confusable ASR slips & 74.2 / 5.3 & 79.6 / 5.7 & 80.9 / 6.8 \\
    \bottomrule
  \end{tabular}
\end{table}

\textbf{Clean instructions and fillers.} Every judge commits correctly on 99.8--100\% of clean instructions with no wrong commits, on average at the oracle's earliest word, saving 0.79\,s per instruction. Unseen fillers change almost nothing (99.3--100\%). Keyword matching does as well, so templates alone need no learned model.

\textbf{Unseen words.} Keyword matching never commits wrongly but ignores every unknown word, so it commits on only 30--47\% of instructions with unseen synonyms or slips. Laya with augmentation commits correctly on 56--77\%, which is 1.6 to 2 times as often, with 3.9--5.0\% wrong commits. When it commits it is right 92--95\% of the time, and even the lower 95\% bound stays above the 90\% set by $\tau$ (90.2--93.6\%). Its correct-commit intervals never overlap with those of keyword matching. Its commits save 0.79--1.06\,s per instruction, within 0.05\,s of the earliest word that determines the target. It commits on 3.6--4.8\% of ambiguous instructions. Laya without augmentation is more conservative. It commits correctly on 39--61\% with 1.6--3.8\% wrong commits and at most 1.2\% false commits. On confusable slips all judges err on 5--7\% because the heard word names a different object.

\textbf{The threshold as a dial.} Figure~\ref{fig:tradeoff} shows the trade-off as $\tau$ changes. Lowering $\tau$ buys more correct commits at the price of more wrong ones. Above $\tau\approx0.85$ the curves stop moving. Raising $\tau$ from 0.85 to 0.95 changes no decision, because the first prefix that passes 0.85 also passes 0.95. On all combined unseen language, prefixes with a top probability above 0.9 average 0.98 but are right only 92--95\% of the time. For Laya with augmentation this sets a floor of about 4--5\% wrong commits on unseen language that $\tau$ alone cannot remove.


\textbf{Calibrate on words the model has not seen.} Table~\ref{tab:calib} fits the temperature on the same word pool used for augmentation. The model already knows these words, so the fit sharpens its probabilities ($T=0.61$) and looks better on familiar language. On unseen language it triples wrong commits, nearly quadruples false commits, and accuracy at commit falls to 0.81, with an upper 95\% bound of 0.83, clearly below the 0.9 promised by $\tau$. Its probabilities between 0.65 and 0.86 are right only 34--44\% of the time, and its prefixes above 0.9 claim 0.997 on average but are right 82\% of the time. It also commits before the words decide. Its commits save 1.18\,s where the earliest determining word allows only 1.09\,s, so part of its extra coverage is guessing. Fitting on the disjoint calibration pool softens the probabilities ($T=5.29$) and restores the margin at a cost of 6 points of correct commits. This follows from the assumption that calibration data resembles deployment data~\cite{guo2017calibration,ovadia2019trust,angelopoulos2025ltt}. We found no prior work stating it for augmented training, and it is easy to miss because the bad fit looks better on familiar data.

\textbf{Speed.} Across repeated runs on a laptop RTX 3060 and a Colab T4 GPU, the fine-tuned judge takes 31--35\,ms per word at the median and 38--41\,ms at the 95th percentile, well within the roughly 300\,ms it takes to speak a word.

\begin{table}[t]
  \caption{Where the temperature is fitted matters. Laya with augmentation, $\tau=0.9$, all combined unseen language unless stated.}
  \label{tab:calib}
  \small
  \begin{tabular}{@{}lcc@{}}
    \toprule
    Temperature fitted on & Training pool & Calibration pool \\
    \midrule
    Fitted temperature $T$ & 0.61 & 5.29 \\
    Calibration error, clean & 0.001 & 0.009 \\
    Calibration error, unseen & 0.070 & 0.026 \\
    Correct commits & 65.0\% & 59.0\% \\
    Wrong commits & 15.2\% & 5.0\% \\
    Accuracy at commit & 0.81 & 0.92 \\
    False commits, ambiguous & 15.2\% & 4.0\% \\
    \bottomrule
  \end{tabular}
\end{table}
\section{Code and Software}

\tool{} is a Python package of 14 modules (\texttt{src/earlycommit}) covering data generation, the judge and baselines, calibration, the commit rule, training, evaluation, streaming speech and the demo. \texttt{python -m earlycommit.paper} regenerates every table, figure and headline number from the included results on a CPU, and three Colab notebooks cover training, evaluation and the demo. All randomness is seeded. It was tested on Ubuntu (Python~3.10) and Colab, needs \texttt{laya}~$\geq$0.3.7 (plus \texttt{faster-whisper}, gTTS and FFmpeg for speech), and is released under Apache~2.0.

\section{Usage Notes}

\textbf{Using your own robot.} Pass the object list from your perception system and the word stream from any recogniser, and hand the returned object to your grasp controller. New objects need no retraining, but the attribute vocabulary is that of the templates. Other vocabularies need new templates and a new fine-tune.

\textbf{Choosing $\tau$.} Read $1-\tau$ as the accepted chance of choosing the wrong object. In our setting it behaves as intended up to about 0.85. For stricter guarantees, replace the threshold with a risk-control method such as Learn then Test~\cite{angelopoulos2025ltt,ringel2024early}.

\textbf{Calibrating.} Fit the temperature on language variation that was not used in training, and refit it when moving to new speakers or vocabularies.

\textbf{Limitations and responsible use.} Instructions are templated and cover a small tabletop vocabulary. The evaluation uses text with simulated slips, and speech is only exercised in the demo with text-to-speech or a recording. There is no user study and no physical robot. The threshold is a calibrated heuristic without a finite-sample guarantee, and a commit is never revised. A robot that moves while a person speaks must keep its usual safety layers, and $\tau$ is not a safety certification. This work involves no human participants.

\textbf{Use of generative AI.}An AI assistant helped correct grammar, and the authors reviewed every change. The authors designed the tool and experiments, reviewed all text and take full responsibility for the content.

\balance
\bibliographystyle{ACM-Reference-Format}
\bibliography{software}

@String{Computing = "Computing" }

@String{Computer = "{IEEE} Computer" }

@inproceedings{schlangen2009incremental,
    title = "Incremental Reference Resolution: The Task, Metrics for Evaluation, and a {B}ayesian Filtering Model that is Sensitive to Disfluencies",
    author = "Schlangen, David  and
      Baumann, Timo  and
      Atterer, Michaela",
    editor = "Healey, Patrick  and
      Pieraccini, Roberto  and
      Byron, Donna  and
      Young, Steve  and
      Purver, Matthew",
    booktitle = "Proceedings of the {SIGDIAL} 2009 Conference",
    month = sep,
    year = "2009",
    address = "London, UK",
    publisher = "Association for Computational Linguistics",
    url = "https://aclanthology.org/W09-3905/",
    pages = "30--37"
}

@inproceedings{devault2009finish,
    title = "Can {I} Finish? Learning When to Respond to Incremental Interpretation Results in Interactive Dialogue",
    author = "DeVault, David  and
      Sagae, Kenji  and
      Traum, David",
    editor = "Healey, Patrick  and
      Pieraccini, Roberto  and
      Byron, Donna  and
      Young, Steve  and
      Purver, Matthew",
    booktitle = "Proceedings of the {SIGDIAL} 2009 Conference",
    month = sep,
    year = "2009",
    address = "London, UK",
    publisher = "Association for Computational Linguistics",
    url = "https://aclanthology.org/W09-3902/",
    pages = "11--20"
}

@inproceedings{kennington2015simple,
    title = "Simple Learning and Compositional Application of Perceptually Grounded Word Meanings for Incremental Reference Resolution",
    author = "Kennington, Casey  and
      Schlangen, David",
    editor = "Zong, Chengqing  and
      Strube, Michael",
    booktitle = "Proceedings of the 53rd Annual Meeting of the Association for Computational Linguistics and the 7th International Joint Conference on Natural Language Processing (Volume 1: Long Papers)",
    month = jul,
    year = "2015",
    address = "Beijing, China",
    publisher = "Association for Computational Linguistics",
    url = "https://aclanthology.org/P15-1029/",
    doi = "10.3115/v1/P15-1029",
    pages = "292--301"
}

@inproceedings{madureira2020incremental,
    title = "Incremental Processing in the Age of Non-Incremental Encoders: An Empirical Assessment of Bidirectional Models for Incremental {NLU}",
    author = "Madureira, Brielen  and
      Schlangen, David",
    editor = "Webber, Bonnie  and
      Cohn, Trevor  and
      He, Yulan  and
      Liu, Yang",
    booktitle = "Proceedings of the 2020 Conference on Empirical Methods in Natural Language Processing (EMNLP)",
    month = nov,
    year = "2020",
    address = "Online",
    publisher = "Association for Computational Linguistics",
    url = "https://aclanthology.org/2020.emnlp-main.26/",
    doi = "10.18653/v1/2020.emnlp-main.26",
    pages = "357--374"
}

@inproceedings{kennington2025prior,
    title = "Prior Lessons of Incremental Dialogue and Robot Action Management for the Age of Language Models",
    author = "Kennington, Casey  and
      Lison, Pierre  and
      Schlangen, David",
    editor = "Buschmeier, Hendrik  and
      Eugenio, Barbara Di  and
      Healey, Patrick  and
      Kennington, Casey  and
      Schlangen, David  and
      Stede, Manfred  and
      Zeldes, Amir",
    booktitle = "Dialogue \& Discourse Volume 16",
    month = "12",
    year = "2025",
    url = "https://aclanthology.org/2025.dnd-16.12/",
    doi = "10.5210/dad.2025.305",
    pages = "96--130",
  
}

@inproceedings{brick2007incremental,
author = {Brick, Timothy and Scheutz, Matthias},
title = {Incremental natural language processing for HRI},
year = {2007},
isbn = {9781595936172},
publisher = {Association for Computing Machinery},
address = {New York, NY, USA},
url = {https://doi.org/10.1145/1228716.1228752},
doi = {10.1145/1228716.1228752},
booktitle = {Proceedings of the ACM/IEEE International Conference on Human-Robot Interaction},
pages = {263–270},
numpages = {8},
location = {Arlington, Virginia, USA},
series = {HRI '07}
}

@inproceedings{hough2017grounding,
author = {Hough, Julian and Schlangen, David},
title = {It's Not What You Do, It's How You Do It: Grounding Uncertainty for a Simple Robot},
year = {2017},
isbn = {9781450343367},
publisher = {Association for Computing Machinery},
address = {New York, NY, USA},
url = {https://doi.org/10.1145/2909824.3020214},
doi = {10.1145/2909824.3020214},
booktitle = {Proceedings of the 2017 ACM/IEEE International Conference on Human-Robot Interaction},
pages = {274–282},
numpages = {9},
location = {Vienna, Austria},
series = {HRI '17}
}

@inproceedings{huang2016anticipatory,
author = {Huang, Chien-Ming and Mutlu, Bilge},
title = {Anticipatory Robot Control for Efficient Human-Robot Collaboration},
year = {2016},
isbn = {9781467383707},
publisher = {IEEE Press},
booktitle = {The Eleventh ACM/IEEE International Conference on Human Robot Interaction},
pages = {83–90},
numpages = {8},
location = {Christchurch, New Zealand},
series = {HRI '16}
}

@inproceedings{shiwa2008quickly,
author = {Shiwa, Toshiyuki and Kanda, Takayuki and Imai, Michita and Ishiguro, Hiroshi and Hagita, Norihiro},
title = {How quickly should communication robots respond?},
year = {2008},
isbn = {9781605580173},
publisher = {Association for Computing Machinery},
address = {New York, NY, USA},
url = {https://doi.org/10.1145/1349822.1349843},
doi = {10.1145/1349822.1349843},
booktitle = {Proceedings of the 3rd ACM/IEEE International Conference on Human Robot Interaction},
pages = {153–160},
numpages = {8},
location = {Amsterdam, The Netherlands},
series = {HRI '08}
}

@misc{park2026premover,
      title={Premover: Fast Vision-Language-Action Control by Acting Before Instructions Are Complete}, 
      author={Joonha Park and Jiseung Jeong and Taesik Gong},
      year={2026},
      eprint={2605.12160},
      archivePrefix={arXiv},
    
}

@inproceedings{ren2023robots,
    title={Robots That Ask For Help: Uncertainty Alignment for Large Language Model Planners},
    author={Ren, Allen Z. and Dixit, Anushri and Bodrova, Alexandra and Singh, Sumeet and Tu, Stephen and Brown, Noah and Xu, Peng and Takayama, Leila and Xia, Fei and Varley, Jake and Xu, Zhenjia and Sadigh, Dorsa and Zeng, Andy and Majumdar, Anirudha},
    booktitle={Proceedings of the Conference on Robot Learning (CoRL)},
    year={2023}
}

@inproceedings{ringel2024early,
author = {Ringel, Liran and Cohen, Regev and Freedman, Daniel and Elad, Michael and Romano, Yaniv},
title = {Early time classification with accumulated accuracy gap control},
year = {2024},
publisher = {JMLR.org},
booktitle = {Proceedings of the 41st International Conference on Machine Learning},
articleno = {1732},
numpages = {17},
location = {Vienna, Austria},
series = {ICML'24}
}

@article{angelopoulos2025ltt,
  title={Learn then test: Calibrating predictive algorithms to achieve risk control},
  author={Angelopoulos, Anastasios N and Bates, Stephen and Cand{\`e}s, Emmanuel J and Jordan, Michael I and Lei, Lihua},
  journal={The Annals of Applied Statistics},
  volume={19},
  number={2},
  pages={1641--1662},
  year={2025},
  publisher={Institute of Mathematical Statistics}
}

@inproceedings{guo2017calibration,
author = {Guo, Chuan and Pleiss, Geoff and Sun, Yu and Weinberger, Kilian Q.},
title = {On calibration of modern neural networks},
year = {2017},
publisher = {JMLR.org},
booktitle = {Proceedings of the 34th International Conference on Machine Learning - Volume 70},
pages = {1321–1330},
numpages = {10},
location = {Sydney, NSW, Australia},
series = {ICML'17}
}

@inproceedings{ovadia2019trust,
  author    = {Ovadia, Yaniv and Fertig, Emily and Ren, Jie and Nado, Zachary and Sculley, D. and Nowozin, Sebastian and Dillon, Joshua V. and Lakshminarayanan, Balaji and Snoek, Jasper},
  title     = {Can You Trust Your Model's Uncertainty? Evaluating Predictive Uncertainty Under Dataset Shift},
  booktitle = {Proceedings of the 33rd International Conference on Neural Information Processing Systems},
  pages     = {14003--14014},
  year      = {2019}
  
}

@inproceedings{machacek2023turning,
    title = "Turning Whisper into Real-Time Transcription System",
    author = "Mach{\'a}{\v{c}}ek, Dominik  and
      Dabre, Raj  and
      Bojar, Ond{\v{r}}ej",
    editor = "Saha, Sriparna  and
      Sujaini, Herry",
    booktitle = "Proceedings of the 13th International Joint Conference on Natural Language Processing and the 3rd Conference of the Asia-Pacific Chapter of the Association for Computational Linguistics: System Demonstrations",
    month = nov,
    year = "2023",
    address = "Bali, Indonesia",
    publisher = "Association for Computational Linguistics",
    url = "https://aclanthology.org/2023.ijcnlp-demo.3/",
    doi = "10.18653/v1/2023.ijcnlp-demo.3",
    pages = "17--24"
}

@inproceedings{yin2019benchmarking,
    title = "Benchmarking Zero-shot Text Classification: Datasets, Evaluation and Entailment Approach",
    author = "Yin, Wenpeng  and
      Hay, Jamaal  and
      Roth, Dan",
    editor = "Inui, Kentaro  and
      Jiang, Jing  and
      Ng, Vincent  and
      Wan, Xiaojun",
    booktitle = "Proceedings of the 2019 Conference on Empirical Methods in Natural Language Processing and the 9th International Joint Conference on Natural Language Processing (EMNLP-IJCNLP)",
    month = nov,
    year = "2019",
    address = "Hong Kong, China",
    publisher = "Association for Computational Linguistics",
    url = "https://aclanthology.org/D19-1404/",
    doi = "10.18653/v1/D19-1404",
    pages = "3914--3923"
}

@inproceedings{warner2024modernbert,
    title = "Smarter, Better, Faster, Longer: A Modern Bidirectional Encoder for Fast, Memory Efficient, and Long Context Finetuning and Inference",
    author = {Warner, Benjamin  and
      Chaffin, Antoine  and
      Clavi{\'e}, Benjamin  and
      Weller, Orion  and
      Hallstr{\"o}m, Oskar  and
      Taghadouini, Said  and
      Gallagher, Alexis  and
      Biswas, Raja  and
      Ladhak, Faisal  and
      Aarsen, Tom  and
      Adams, Griffin Thomas  and
      Howard, Jeremy  and
      Poli, Iacopo},
    editor = "Che, Wanxiang  and
      Nabende, Joyce  and
      Shutova, Ekaterina  and
      Pilehvar, Mohammad Taher",
    booktitle = "Proceedings of the 63rd Annual Meeting of the Association for Computational Linguistics (Volume 1: Long Papers)",
    month = jul,
    year = "2025",
    address = "Vienna, Austria",
    publisher = "Association for Computational Linguistics",
    url = "https://aclanthology.org/2025.acl-long.127/",
    doi = "10.18653/v1/2025.acl-long.127",
    pages = "2526--2547",
    ISBN = "979-8-89176-251-0"
}

@misc{laya2026,
  author       = {{Convai Innovations}},
  title        = {Laya (model card)},
  year         = {2026},
  howpublished = {\url{https://huggingface.co/convaiinnovations/laya}}
}

@inproceedings{radford2023whisper,
author = {Radford, Alec and Kim, Jong Wook and Xu, Tao and Brockman, Greg and McLeavey, Christine and Sutskever, Ilya},
title = {Robust speech recognition via large-scale weak supervision},
year = {2023},
publisher = {JMLR.org},
booktitle = {Proceedings of the 40th International Conference on Machine Learning},
articleno = {1182},
numpages = {27},
location = {Honolulu, Hawaii, USA},
series = {ICML'23}
}

@article{SKANTZE2021101178,
  author  = {Skantze, Gabriel},
  title   = {Turn-taking in Conversational Systems and Human-Robot Interaction: A Review},
  journal = {Computer Speech \& Language},
  volume  = {67},
  pages   = {101178},
  year    = {2021},
  doi     = {10.1016/j.csl.2020.101178}
}

\end{document}